\documentclass{article}
\usepackage{spconf,amsmath,graphicx}
\usepackage{amssymb} 
\usepackage{epsfig}

\title{Elastic Registration of Medical Images With GANs}
\name{Dwarikanath Mahapatra}
\address{IBM Research - Australia, Melbourne}
\begin{document}
%
\maketitle
\begin{abstract}
Conventional approaches to image registration consist of time consuming iterative methods. Most current deep learning (DL) based registration methods extract deep features to use in an iterative setting. We propose an end-to-end DL method for registering multimodal images. Our approach uses generative adversarial networks (GANs) that eliminates the need for time consuming iterative methods, and directly generates the registered image with the deformation field. Appropriate constraints in the GAN cost function produce accurately registered images in less than a second. Experiments demonstrate their accuracy for multimodal retinal and cardiac MR image registration.
\end{abstract}
%


\section{Introduction}
\label{sec:Intro}

Image registration is a fundamental step in most medical image analysis problems, and a comprehensive review of algorithms can be found in \cite{Reg_review1,Mahapatra_CVIU_2019,Mahapatra_CMIG2019,Mahapatra_LME_PR2017,
Zilly_CMIG_2016}. Conventional registration methods use iterative gradient descent based optimization using cost functions such as mean square error (MSE), normalized mutual information, etc. Such methods tend to be time consuming, especially for volumetric images. We propose a fully end-to-end deep learning (DL) approach that does not employ iterative methods, but uses generative adversarial networks (GANs) for obtaining  registered images and the corresponding deformation field.

Thw works in \cite{WuTBME,Mahapatra_SSLAL_CD_CMPB,Mahapatra_SSLAL_Pro_JMI,Mahapatra_LME_CVIU,LiTMI_2015} use convolutional stacked autoencoders (CAE) to extract features from fixed and moving images, and use it in a conventional iterative deformable registration framework.
The works of  \cite{Miao_Reg,MahapatraJDI_Cardiac_FSL,Mahapatra_JSTSP2014,MahapatraTIP_RF2014,MahapatraTBME_Pro2014,MahapatraTMI_CD2013} use convolutional neural network (CNN) regressors in rigid registration of synthetic images. In \cite{Liao_Reg,MahapatraJDICD2013,MahapatraJDIMutCont2013,MahapatraJDIGCSP2013,MahapatraJDIJSGR2013} employ CNNs and reinforcement learning for iterative registration of CT to cone-beam CT in cardiac and abdominal images. 
DL based regression methods still require  conventional methods to generate the transformed image. 

Jaderberg et al. \cite{STN} introduced spatial transformer networks (STN) to align input images in a larger task-specific network. STNs, however, need many labeled training examples and have not been used for medical image analysis. 
%
%
Sokooti et. al. \cite{RegNet,MahapatraJDISkull2012,MahapatraTIP2012,MahapatraTBME2011,MahapatraEURASIP2010} propose RegNet that uses CNNs trained on simulated deformations to generate displacement vector fields for a pair of unimodal images. Vos et. al. \cite{Vos_DIR,Bozorgtabar_ICCV19,Xing_MICCAI19,Mahapatra_ISBI19,MahapatraAL_MICCAI18,Mahapatra_MLMI18} propose the deformable image registration network (DIR-Net) which takes pairs of fixed and moving images as input, and outputs a transformed image non-iteratively. Training is completely unsupervised and unlike previous methods it is not trained with known registration transformations. 

While RegNet and DIRNet are among the first methods to achieve registration in a single pass, they have some limitations such as: 1) using spatially corresponding patches to predict transformations. Finding corresponding patches is challenging in low contrast medical images and can adversely affect the registration task; 2) Multimodal registration is challenging with their approach due to the inherent problems of finding spatially corresponding patches; 3) DIRNet uses B-splines for spatial transformations which limits the extent of recovering a deformation field; 4) Use of intensity based cost functions limits the benefits that can be derived from a DL based image registration framework.

To overcome the above limitations we make the following contributions: 1) we use GANs for multimodal medical image registration, which can recover more complex range of deformations ; 2) novel constraints in the cost function, such as VGG, SSIM loss and deformation field reversibility, ensure that the trained network can easily generate images that are realistic with a plausible deformation field. We can choose any image as the reference image and registration is achieved in a single pass.



\section{Methods}
\label{sec:met}

GANs are generative DL models trained to output many image types. Training is performed in an adversarial setting where a discriminator outputs a probability of the generated image matching the training data distribution. GANs have been used in various applications such as image super resolution \cite{SRGAN,MahapatraMICCAI_ISR,Sedai_OMIA18,Sedai_MLMI18,MahapatraGAN_ISBI18,Sedai_MICCAI17,Mahapatra_MICCAI17}, image synthesis and image translation using conditional GANs (cGANs) \cite{CondGANs,Roy_ISBI17,Roy_DICTA16,Tennakoon_OMIA16,Sedai_OMIA16,Mahapatra_MLMI16} and cyclic GANs (cycGANs) \cite{CyclicGANs,Sedai_EMBC16,Mahapatra_EMBC16,Mahapatra_MLMI15_Optic,Mahapatra_MLMI15_Prostate,Mahapatra_OMIA15}. 
  
In cGANs the output is conditioned on the input image and a random noise vector, and requires training image pairs. On the other hand cycGANs do not require training image pairs but enforce consistency of deformation field. We leverage the advantage of both methods to register multimodal images.
%
%
For multimodal registration we use cGANs to ensure the generated output image (i.e., the transformed floating image) has the same characteristic as the floating image (in terms of intensity distribution) while being similar to the reference image (of a different modality) in terms of landmark locations. This is achieved by incorporating appropriate terms in the loss function for image generation. Additionally, we enforce deformation consistency to obtain realistic deformation fields. This prevents unrealistic registrations and allows any image to be the reference or floating image. A new test image pair from modalities not part of the training set can be registered without the need for re-training the network.

\subsection{Generating Registered Images}

Let us denote the registered (or transformed) image as $I^{Trans}$, obtained from the input floating image $I^{Flt}$, and is to be registered to the fixed reference image $I^{Ref}$. For training we have pairs of multimodal images where the corresponding landmarks are perfectly aligned (e.g., retinal fundus and fluoroscein angiography (FA) images). Any one of the modalities (say fundus) is $I^{Ref}$. $I^{Flt}$ is generated by applying a known elastic deformation field to the other image modality (in this case FA). The goal of registration is to obtain $I^{Trans}$ from $I^{Flt}$ such that $I^{Trans}$ is aligned with $I^{Ref}$. Applying synthetic deformations allows us to: 1) accurately quantify the registration error in terms of deformation field recovery; and 2) determine the similarity between $I^{Trans}$ and FA images.

The generator network that outputs $I^{Trans}$ from $I^{Flt}$ is a feed-forward CNN whose parameters $\theta_G$ are, 
\begin{equation}
\widehat{\theta}=\arg \min_{\theta_G} \frac{1}{N} \sum_{n=1}^{N} l^{SR}\left(G_{\theta_G}(I^{Flt}),I^{Ref},I^{Flt}\right),
\label{eq:theta1}
\end{equation}
where the loss function $l^{SR}$ combines content loss (to ensure that $I^{Trans}$ has desired characteristics) and adversarial loss, and $G_{\theta_G}(I^{Flt})=I^{Trans}$. The content loss is, 
\begin{equation}
\begin{split}
l_{content}& = NMI(I^{Trans},I^{Ref}) + SSIM(I^{Trans},I^{Ref}) \\ 
 & + VGG(I^{Trans},I^{Ref}).
\end{split}
\label{eq:conLoss}
\end{equation} 

$I^{Trans}$ should: 1) have identical intensity distribution as $I^{Flt}$ and; 2) have similar structural information content as $I^{Ref}$.
$NMI(I^{Trans},I^{Ref})$ denotes the normalized mutual information (NMI) between $I^{Ref}$ and $I^{Trans}$. NMI is a widely used cost function for multimodal deformable registration \cite{FFD,MahapatraISBI15_Optic,MahapatraISBI15_JSGR,MahapatraISBI15_CD,KuangAMM14,Mahapatra_ABD2014} since it matches the joint intensity distribution of two images. 
$SSIM(I^{Trans},I^{Ref})$ denotes the structural similarity index metric (SSIM) \cite{SSIM,Schuffler_ABD2014,MahapatraISBI_CD2014,MahapatraMICCAI_CD2013,Schuffler_ABD2013,MahapatraProISBI13} and calculates image similarity based on edge distribution and other landmarks. Since it is not based on intensity values it accurately quantifies landmark correspondence between different images. $VGG(I^{Trans},I^{Ref})$ is the $L2$ distance between two images using all $512$ feature maps of Relu $4-1$ layer of a pre-trained $VGG-16$ network \cite{VGG,MahapatraRVISBI13,MahapatraWssISBI13,MahapatraCDFssISBI13,MahapatraCDSPIE13,MahapatraABD12}. VGG loss improves robustness since the cost function takes into account multiple feature maps that capture information at different scales. 

The adversarial loss of Eqn.~\ref{eqn:cyGan1} ensures that $I^{Trans}$ has an identical intensity distribution as $I^{Flt}$. We could realize this condition by having an extra $NMI(I^{Trans},I^{Flt})$ term in Eqn.~\ref{eq:conLoss}. But this does not lead to much improvement in results.

\begin{figure}[t]
\begin{tabular}{c}
\includegraphics[height=2.9cm, width=7.99cm]{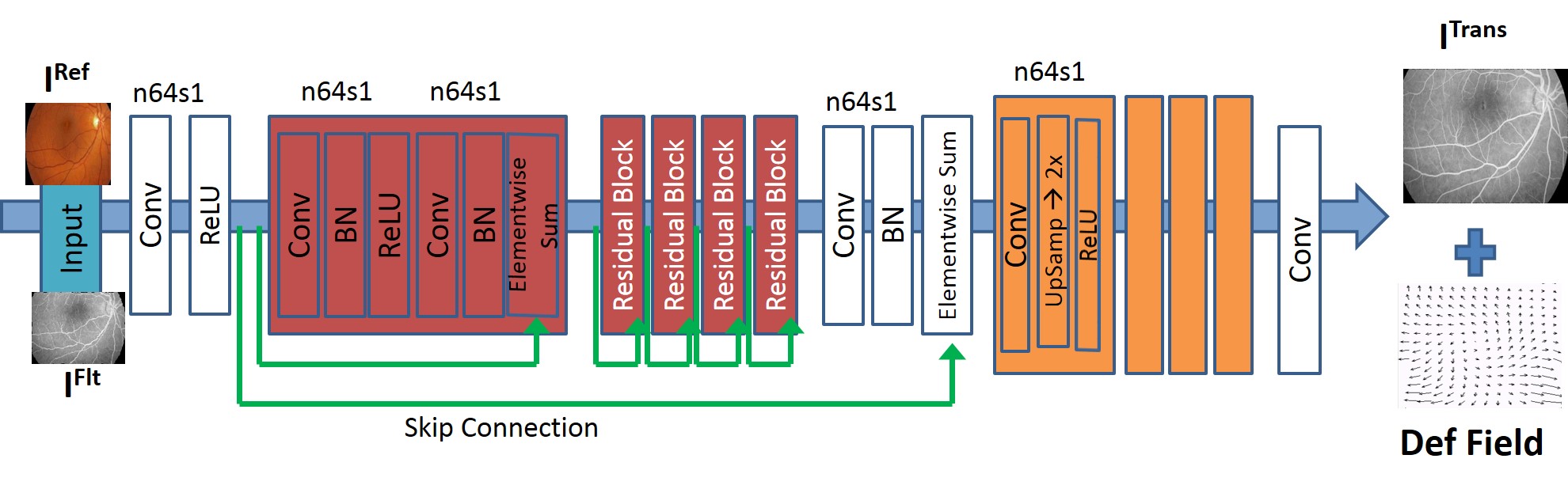}  \\
(a) \\
\includegraphics[height=2.9cm, width=7.99cm]{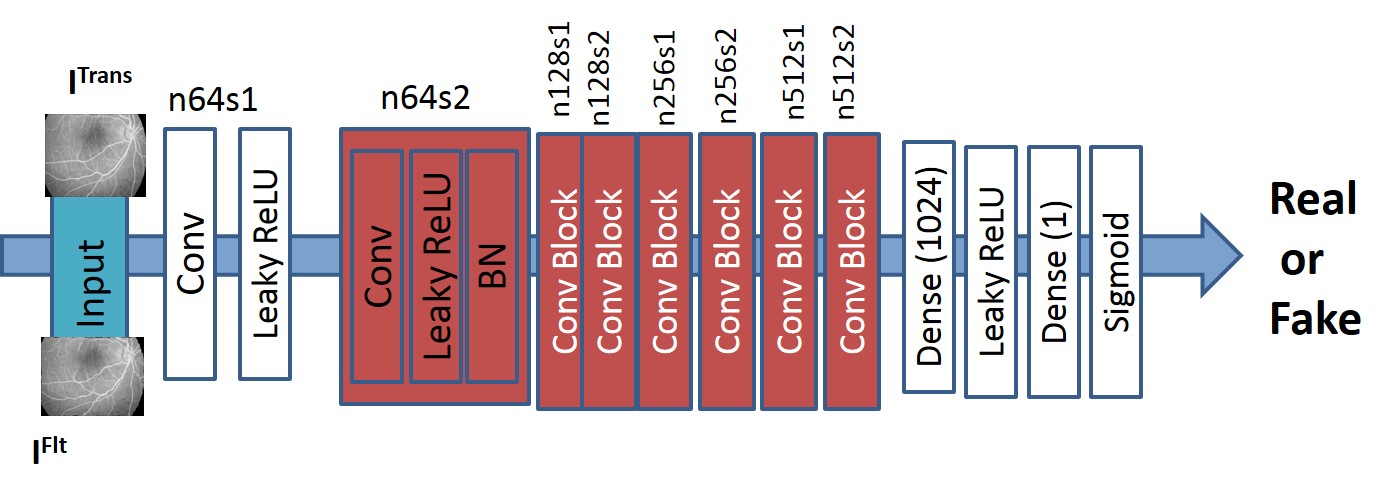}  \\
(b)\\
\end{tabular}
\caption{(a) Generator Network; (b) Discriminator network. $n64s1$ denotes $64$ feature maps (n) and stride (s) $1$ for each convolutional layer.}
\label{fig:Gan}
\end{figure}

The generator network $G$ (Figure~\ref{fig:Gan}(a)) employs residual blocks, each block having two convolutional layers with $3\times3$ filters and $64$ feature maps, followed by batch normalization and ReLU activation. In addition to generating the registered image $G$ also outputs a deformation field. 
The discriminator $D$ (Figure~\ref{fig:Gan} (b)) has eight convolutional layers with the kernels increasing by a factor of $2$ from $64$ to $512$ . Leaky ReLU is used and strided convolutions reduce the image dimension  when the number of features is doubled. The resulting $512$ feature maps are
followed by two dense layers and a final sigmoid activation to obtain a probability map. $D$ evaluates similarity of intensity distribution between $I^{Trans}$ and $I^{Ref}$, and the error between generated and reference deformation fields.

\subsection{Deformation Field Consistency}


CycGANs learn mapping functions between two domains
$X$ and $Y$ given training samples $x_{i=1}^{N} \in X$ and $y_{j=1}^{M} \in Y$. It has two transformations $G : X \rightarrow Y$ and $F : Y \rightarrow X$, and two adversarial discriminators $D_X$ and $D_Y$ , where $D_X$ differentiates  between images ${x}$ and registered images ${F(y)}$ and $D_Y$ distinguishes between ${y}$ and ${G(x)}$. Here $X=I^{Flt}$ and $Y=I^{Ref}$. $G$ registers $I^{Flt}$ to $I^{Ref}$ while $F$ registers $I^{Ref}$ to $I^{Flt}$.
In addition to the content loss (Eqn~\ref{eq:conLoss}) we have: 1) an adversarial loss to match $I^{Trans}$'s distribution to $I^{Flt}$; and 2) a cycle consistency loss to ensure transformations $G,F$ do not contradict each other.

\subsubsection{Adversarial Loss}

The adversarial loss function for $G$ is given by:
\begin{equation}
\begin{split}
& L_{cycGAN}(G,D_Y,X,Y) = E_{y\in p_{data}(y)} \left[\log D_Y(y)\right] + \\ 
& E_{x\in p_{data}(x)} \left[\log \left(1-D_Y(G(x))\right)\right],  \\
\end{split}
\label{eqn:cyGan1}
\end{equation}
We retain notations $X,Y$ for conciseness.
%
 There also exists $L_{cycGAN}(F,D_X,Y,X)$ the corresponding adversarial loss for $F$ and $D_X$. 

\subsubsection{Cycle Consistency Loss}

A network may arbitrarily transform the input image to match the distribution of the target domain. Cycle consistency loss ensures that for each image $x \in X$ the reverse deformation should bring $x$ back to the original image, i.e. $x \rightarrow G(x) \rightarrow F(G(x))\approx x$. 
Similar constraints also apply for mapping $F$ and $y$. This is achieved using, 
%
%
\begin{equation}
L_{cyc}(G,F)= E_{x} \left\|F(G(x))-x\right\|_1 + E_{y} \left\|G(F(y))-y\right\|_1,
\label{eqn:cyGan2}
\end{equation}
%

The full objective function is 
\begin{equation}
\begin{split}
& L(G,F,D_{I^{Flt}},D_{I^{Ref}})= L_{cycGAN}(G,D_{I^{Ref}},I^{Flt},I^{Ref}) \\ 
& + L_{cycGAN}(F,D_{I^{Flt}},I^{Ref},I^{Flt}) + \lambda L_{cyc}(G,F)
\end{split}
\label{eqn:cyGan3}
\end{equation}
where $\lambda=10$ controls the contribution of the two objectives. The optimal parameters are given by:
\begin{equation}
G^{*},F^{*}=\arg \min_{F,G} \max_{D_{I^{Flt}},D_{I^{Ref}}} L(G,F,D_{I^{Flt}},D_{I^{Ref}})
\label{eqn:CyGan4}
\end{equation}

The above formulation ensures $I^{Trans}$ to be similar to $I^{Flt}$ and also match $I^{Ref}$. We do not need to explicitly condition $I^{Trans}$ on $I^{Ref}$ or $I^{Flt}$ as that is implicit in the cost function (Eqns~\ref{eq:conLoss},\ref{eqn:cyGan1}), which allows any pair of multimodal images to be registered even if the modality was not part of the training set.

\section{Experiments and Results}
\label{sec:expts}

We demonstrate the effectiveness of our approach on retinal and cardiac images. Details on dataset and experimental set up are provided later.  
Our method was implemented with Python and TensorFlow (for GANs). For GAN optimization we use Adam \cite{Adam,MahapatraMLMI12,MahapatraSTACOM12,VosEMBC,MahapatraGRSPIE12,MahapatraMiccaiIAHBD11} with $\beta_1=0.93$ and batch normalization. The ResNet was trained with a learning rate of $0.001$ and $10^{5}$ update iterations. MSE based ResNet was used to initialize $G$. The final GAN was trained with $10^{5}$ update iterations at learning rate $10^{-3}$. Training and test was performed on a NVIDIA Tesla K$40$ GPU with $12$ GB RAM. 

\subsection{Retinal Image Registration Results}
\label{expt:retina}


The data consists of retinal colour fundus images and fluorescein angiography (FA) images obtained from $30$ normal subjects. Both images are $576\times720$ pixels and fovea centred \cite{Alipour2014}. 
Registration ground truth was developed using the Insight Toolkit (ITK). 
The Frangi vesselness\cite{Frangi1998,MahapatraMiccai11,MahapatraMiccai10,MahapatraICIP10,MahapatraICDIP10a,MahapatraICDIP10b,MahapatraMiccai08} feature was utilised to find the vasculature, and the maps were aligned using sum of squared differences (SSD). 
Three out of $30$ images could not be aligned due to poor contrast and one FA image was missing, leaving us with a final set of $26$ registered pairs. 
We use the fundus images as $I^{Ref}$ and generate floating images from the FA images by simulating different deformations (using SimpleITK) such as rigid, affine and elastic deformations(maximum displacement of a pixel was $\pm10$ mm. $1500$ sets of deformations were generated for each image pair giving a total of $39000$ image pairs.

Our algorithm's performance was evaluated using average registration error ($Err_{Def}$) between the applied deformation field and the recovered deformation field. 
Before applying simulated deformation the mean Dice overlap of the vasculature between the fundus and FA images across all $26$ patients is $99.2$, which indicates highly accurate alignment. After simulating deformations the individual Dice overlap reduces considerably depending upon the extent of deformation. The Dice value after successful registration is expected to be higher than before registration. We also calculate the $95$ percentile  Hausdorff Distance ($HD_{95}$) and the mean absolute surface distance (MAD) before and after registration. We calculate the mean square error (MSE) between the registered FA image and the original undeformed FA image to quantify their similarity. 
The intensity of both images was normalized to lie in $[0,1]$. Higher values of Dice and lower values of other metrics indicate better registration.
The average training time for the augmented dataset of $39000$ images was $14$ hours. 

\begin{figure*}[t]
\begin{tabular}{cccccccc}
\includegraphics[height=2.1cm,width=1.85cm]{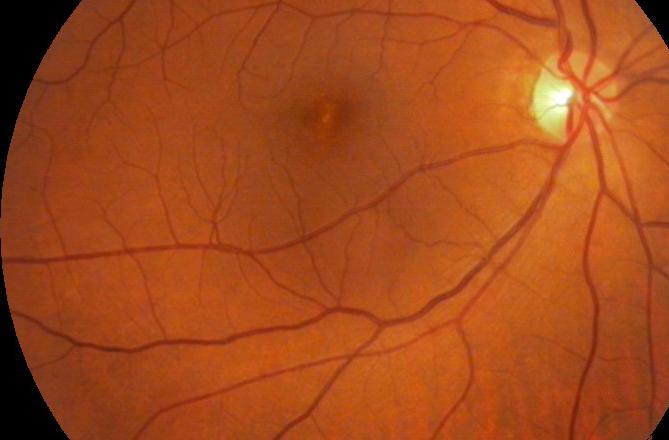} &
\includegraphics[height=2.1cm,width=1.85cm]{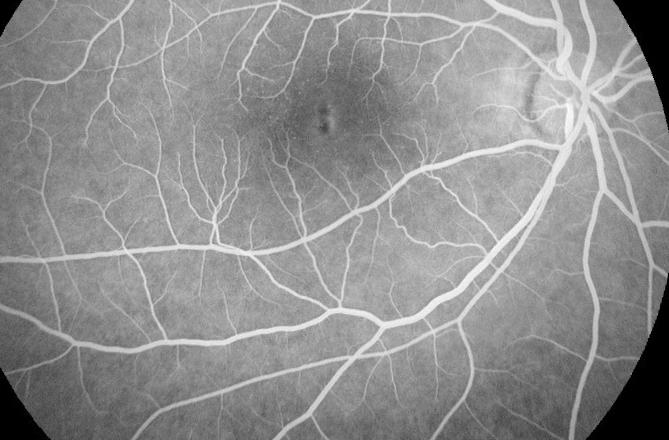} &
\includegraphics[height=2.1cm,width=1.85cm]{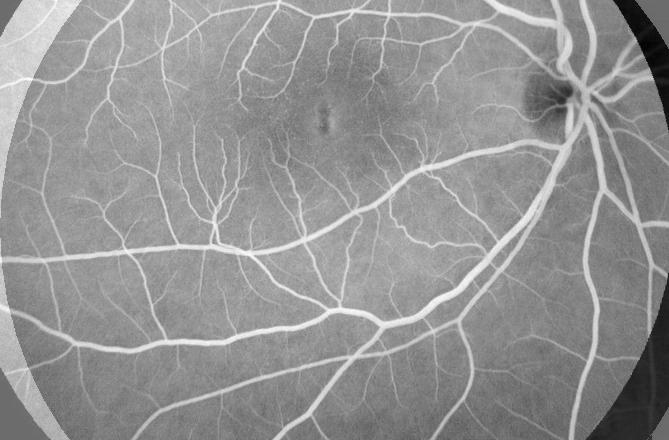} &
\includegraphics[height=2.1cm,width=1.85cm]{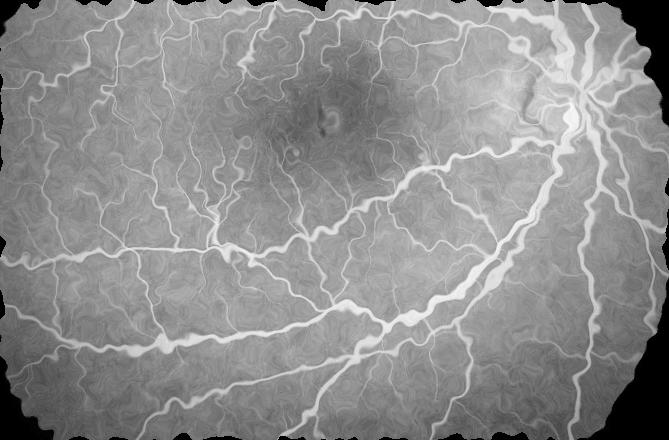} &
\includegraphics[height=2.1cm,width=1.85cm]{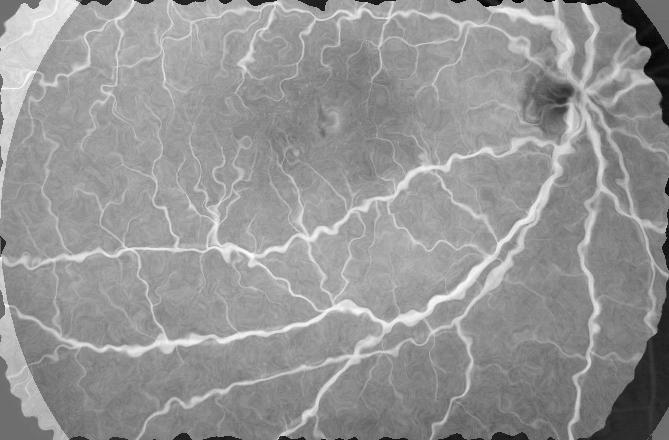} &
\includegraphics[height=2.1cm,width=1.85cm]{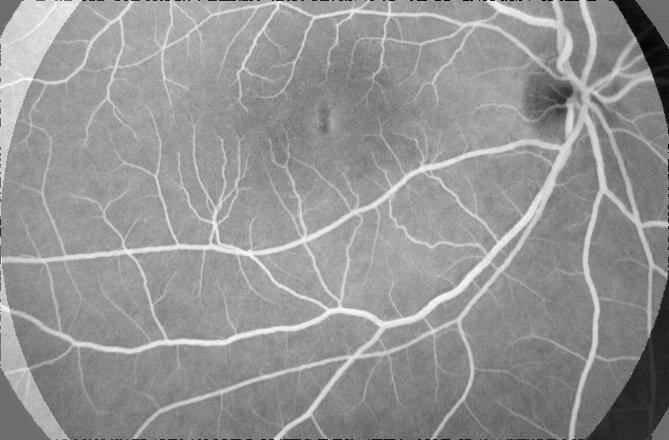} &
\includegraphics[height=2.1cm,width=1.85cm]{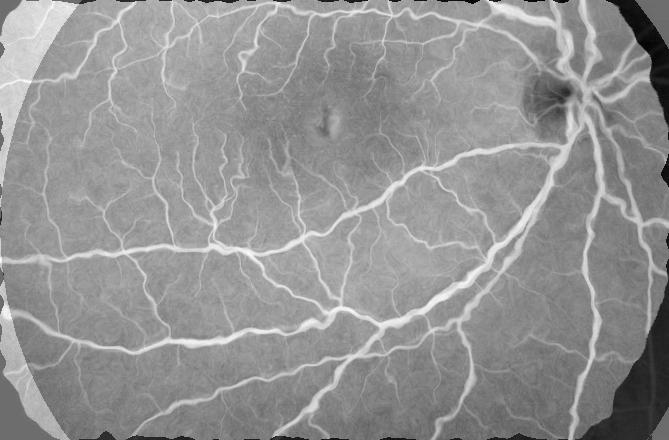} &
\includegraphics[height=2.1cm,width=1.85cm]{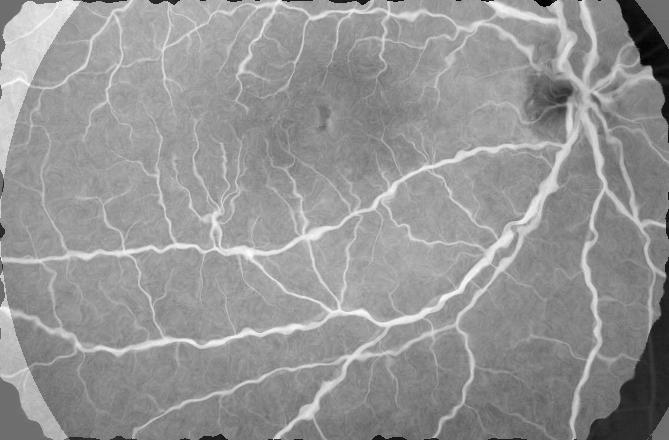} \\
(a) & (b) & (c) & (d) & (e) & (f) & (g) & (h)\\ 
\end{tabular}
\caption{Example results for retinal fundus and FA registration. (a) Color fundus image; (b) Original FA image; (c) ground truth difference image before simulated deformation; (d) Deformed FA image or the floating image; Difference image (e) before registration; after registration using (f) $GAN_{Reg}$; (g) $DIRNet$; (f) Elastix .}
\label{fig:Ret}
\end{figure*}

Table~\ref{tab:Retina} shows the registration performance for $GAN_{Reg}$, our proposed method, and compared with the following methods: $DIRNet$ - the CNN based registration method of \cite{Vos_DIR}; $Elastix$ - an iterative NMI based registration method \cite{Elastix,MahapatraISBI08,MahapatraICME08,MahapatraICBME08_Retrieve,MahapatraICBME08_Sal,MahapatraSPIE08,MahapatraICIT06}; and $GAN_{Reg_{nCyc}}$ - $GAN_{Reg}$ without deformation consistency constraints.  $GAN_{Reg}$ has the best performance across all metrics. 
Figure~\ref{fig:Ret} shows registration results for retinal images. 
$GAN_{Reg}$ registers the images closest to the original and is able to recover most deformations to the blood vessels, followed by $DIRNet$, $GAN_{Reg-nCyc}$, and $Elastix$. It is obvious that deformation reversibility constraints significantly improve registration performance. Note that the fundus images are color while the FA images are grayscale. The reference image is a grayscale version of the fundus image.


\begin{table}[t]
\begin{tabular}{|c|c|c|c|c|c|}
\hline
{}& {} & \multicolumn {4}{|c|}{After Registration} \\  \cline{3-6}
{} & {Bef.} & {$GAN$} & {DIRNet}  & {Elastix} & $GAN$ \\ 
{}& {Reg.} & {$_{Reg}$} & {\cite{Vos_DIR}}  & {\cite{Elastix}} & $_{Reg_{nCyc}}$ \\ \hline
{Dice} & 0.843 & {0.946} & {0.911}  & 0.874 & 0.887\\ \hline
{$Err_{Def}$} & {14.3} & {5.7} & 7.3 & 12.1 & 9.1\\  \hline
{$HD_{95}$}  & {11.4} & {4.2}  & 5.9 & 9.7 & 8.0 \\  \hline
{MAD}   & {9.1} & {3.1}  & 5.0 & 8.7 & 7.2 \\  \hline
{MSE} & {0.84} & {0.09} & {0.23} & {0.54} & 0.37\\ \hline
{$Time$ (s)} & {} & 0.7 & 0.9 & 15.1 & 0.7\\ \hline
\end{tabular}
\caption{Comparative average performance of different methods before and after retinal image registration. $Time$ denotes time in seconds taken to register a test image pair. } 
\label{tab:Retina}
\end{table}


\subsection{Cardiac Image Registration Results}
\label{expt:cardiac}

The second dataset is the Sunybrook cardiac dataset \cite{Sunybrook} with $45$ cardiac cine MRI scans acquired on a single MRI-scanner. They  consist of short-axis cardiac image slices each containing $20$ timepoints that encompass the entire cardiac cycle. Slice thickness and spacing is $8$ mm, and slice dimensions are $256\times256$ with a pixel size of $1.28\times1.28$ mm. The data is equally divided in $15$ training scans ($183$ slices), $15$ validation scans ($168$ slices), and $15$ test scans ($176$ slices). An expert annotated the right ventricle (RV) and left ventricle myocardium at end-diastolic (ED) and end-systolic (ES) time points. Annotations were made in the test scans and only used for final quantitative evaluation. 

We calculate Dice values before and after registration, $HD_{95}$, and MAD. We do not simulate deformations on this dataset and hence do not calculate $Err_{Def},MSE$. Being a public dataset our results can be benchmarked against other methods. While the retinal dataset demonstrates our method's performance for multimodal registration, the cardiac dataset highlights the performance in registering unimodal dynamic images. The network  trained on retinal images was used for registering cardiac data without re-training. The first frame of the sequence was used as the reference image $I^{Ref}$ and all other images were floating images.

\begin{figure}[h]
\begin{tabular}{cccc}
\includegraphics[height=2.0cm,width=1.75cm]{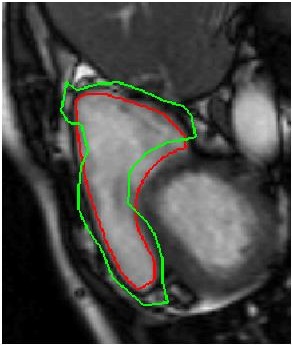} &
\includegraphics[height=2.0cm,width=1.75cm]{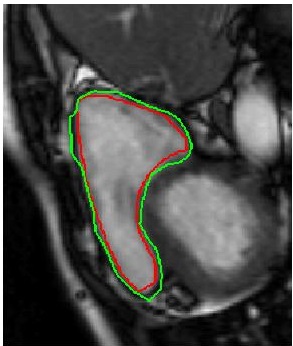} &
\includegraphics[height=2.0cm,width=1.75cm]{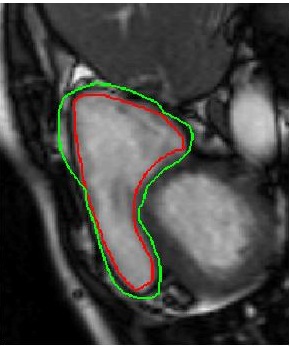} &
\includegraphics[height=2.0cm,width=1.75cm]{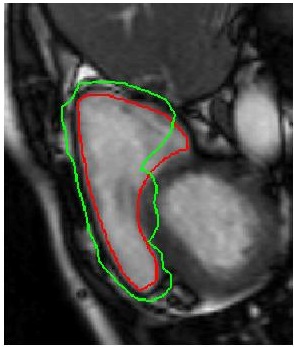} \\
(a) & (b) & (c) & (d)\\ 
\end{tabular}
\caption{Example results for cardiac RV registration. Superimposed contours of the ground truth (red) and deformed segmentation mask of moving image (green): (a) before registration; after registration using (b) $GAN_{Reg}$; (c) $DIRNet$; (d) Elastix.}
\label{fig:Cardiac}
\end{figure}

Table~\ref{tab:cardiac} summarizes the performance of different methods, and Figure~\ref{fig:Cardiac} shows superimposed manual contour of the RV (red) and the deformed contour of the registered image (green). Better registration is reflected by closer alignment of the two contours. Once again it is obvious that $GAN_{Reg}$ has the best performance amongst all competing methods, and its advantages over $GAN_{Reg-nCyc}$ when including deformation consistency.



\begin{table}[t]
\begin{tabular}{|c|c|c|c|c|c|}
\hline
{}& {} & \multicolumn {4}{|c|}{After Registration} \\  \cline{3-6}
{} & {Bef.} & {$GAN$} & {DIRNet}  & {Elastix} & $GAN$\\ 
{}& {Reg.} & {$_{Reg}$} & {\cite{Vos_DIR}}  & {\cite{Elastix}} & $_{Reg_{nCyc}}$\\ \hline
{Dice} & $0.62$ & {$0.85$} & {$0.80$}  &  $0.77$ & 0.79\\ \hline
{$HD_{95}$}  & {$7.79$} & {$3.9$}  & $5.03$ &  $5.21$ & 5.12 \\  \hline
{MAD}   & {$2.89$} & {$1.3$}  & $1.83$ & $2.12$ & 1.98 \\  \hline
{$Time$ (s)} & {} & 0.8 & 0.8 & 11.1 & 0.8 \\ \hline
\end{tabular}
\caption{Comparative average performance of different methods before and after cardiac image registration. $Time$ denotes time in seconds taken to register a test image pair..}
\label{tab:cardiac}
\end{table}

\section{Conclusion}
\label{sec:concl}

We have proposed a GAN based method for multimodal medical image registration. Our proposed method allows fast and accurate registration and is independent of the choice of reference or floating image. Our primary contribution is in using GAN for medical image registration, and combining conditional and cyclic constraints to obtain realistic and smooth registration. Experimental results demonstrate that we perform better than traditional iterative registration methods and other DL based methods that use conventional transformation approaches such as B-splines.

\bibliographystyle{IEEEbib}
\bibliography{ISBI2018_GAN_Ref}

\end{document}